\documentclass{article}

\PassOptionsToPackage{numbers, compress}{natbib}
\usepackage[preprint]{neurips_2024}

\usepackage[utf8]{inputenc} % allow utf-8 input
\usepackage[T1]{fontenc}    % use 8-bit T1 fonts
\usepackage{url}            % simple URL typesetting
\usepackage{booktabs}       % professional-quality tables
\usepackage{amsfonts}       % blackboard math symbols
\usepackage{nicefrac}       % compact symbols for 1/2, etc.
\usepackage{microtype}      % microtypography

\usepackage{graphicx}
\usepackage{algorithm}

\usepackage{wrapfig}
\usepackage[pagebackref=true,breaklinks=true,colorlinks,bookmarks=false]{hyperref}
\usepackage[table,xcdraw]{xcolor}
\usepackage{amsmath}

\usepackage[utf8]{inputenc} % allow utf-8 input
\usepackage{hyperref}       % hyperlinks
\usepackage{subfig}  
\usepackage{caption}

\usepackage{amssymb}
\usepackage{multirow}
\usepackage{amsmath}
\usepackage{cases}
\usepackage{algorithm}
\usepackage{graphicx}
\usepackage{subcaption}
\usepackage{xcolor}

\usepackage{amsfonts}       % blackboard math symbols
\usepackage{nicefrac}       % compact symbols for 1/2, 

\title{MixBCT: Towards Self-Adapting Backward-Compatible Training}

\author{
    Yu Liang\textsuperscript{\rm 1,3}\footnotemark[1]  \quad
    Yufeng Zhang\textsuperscript{\rm 1}  \quad 
    Shiliang Zhang\textsuperscript{\rm 2} \quad
    Yaowei Wang  \textsuperscript{\rm 4} \quad 
    \textbf{Sheng Xiao}\textsuperscript{\rm 1}  \\ 
    \textbf{Rong Xiao}\textsuperscript{\rm 3} \quad   
    \textbf{Xiaoyu Wang}\textsuperscript{\rm 5}\\ 
   \textsuperscript{\rm 1}Hunan University \quad 
    \textsuperscript{\rm 2}Peking University \quad
    \textsuperscript{\rm 3}Intellifusion Inc. \quad
    \textsuperscript{\rm 4}Peng Cheng Laboratory \\
    \textsuperscript{\rm 5}The Hong Kong University of Science and Technology(Guangzhou) 
    \texttt{}
}
\vspace{-1em}

\begin{document}
\renewcommand{\thefootnote}{\fnsymbol{footnote}}
\footnotetext[1]{Work done as an intern at Intellifusion.}

\maketitle


%\input{Abstract}
%\input{Introduction}
%\input{Related_Works}
%\input{Method}
%\input{Expriment}
%\input{Discussion}
%\input{Appendix}

\begin{abstract}
Backward-compatible training circumvents the need for expensive updates to the old gallery database when deploying an advanced new model in the retrieval system. Previous methods achieved backward compatibility by aligning prototypes of the new model with the old one, yet they often overlooked the distribution of old features, limiting their effectiveness when the low quality of the old model results in a weakly feature discriminability. Instance-based methods like L2 regression take into account the distribution of old features but impose strong constraints on the performance of the new model itself. In this paper, we propose MixBCT, a simple yet highly effective backward-compatible training method that serves as a unified framework for old models of varying qualities. We construct a single loss function applied to mixed old and new features to facilitate backward-compatible training, which adaptively adjusts the constraint domain for new features based on the distribution of old features.  We conducted extensive experiments on the large-scale face recognition datasets MS1Mv3 and IJB-C to verify the effectiveness of our method. The experimental results clearly demonstrate its superiority over previous methods. Code is available at https://github.com/yuleung/MixBCT .

\end{abstract}

\section{Introduction}

Image retrieval is a widely-used technique, especially in large-scale industrial settings \cite{horster2007image,jegou2010improving,grauman2013learning,li2018large}. Typically, given a query, the retrieval system returns several related items in the gallery set according to a ranked list of similar entries. With advances in representation learning and increased data availability, there is a natural desire to improve the performance of retrieval systems by training better embedding models. Unfortunately, the embedding produced by the new model may not be compatible with the old embedding, meaning that the old and new embeddings cannot be directly mutually retrieved. Updating old embeddings through an operation called ‘backfilling’ is very costly for large-scale gallery sets consisting of millions or even billions of items. Furthermore, owing to privacy concerns or other issues, the old raw data might no longer be accessible, rendering us unable to update the old gallery database in this scenario. Backward-compatible training, originally proposed in \cite{shen2020BCT}, enables straightforward deployment of new models in a ‘backfill-free’ manner.

BCT \cite{shen2020BCT} achieves backward compatibility by utilizing the classifier of the old model as a regularization term (referred to as the influence loss) during the training of the new model. Specifically, BCT aligns the new model classifier with the class centers of old features by leveraging the old classifier to classify new features.  Recent works \cite{zhang2022uniBCT,bai2022dual,su2022BiCT,zhang2022darwinian,pan2023boundary}, which we refer to as old prototype-based methods, have demonstrated performance enhancement within the BCT framework. 

However, these methods neglect the intra-class distribution information of old features, causing their performance to be severely affected as intra-class distribution variance increases with the deterioration of the old model's quality. In real scenarios, the prevalence of low-quality old gallery features—exacerbated by limitations such as early inferior representation learning techniques, training dataset scale, and other factors—further worsens this issue.

In fact, old prototype-based methods have several limitations beyond their inability to capture the intra-class distribution information of old features. For instance, the old prototypes aligned by the new model may be inaccurate or even erroneous. This inaccuracy can stem from the low quality of the old models themselves. Sometimes, it is necessary to represent old prototypes with class-averaged features of old data when new data contains classes not present in the old data or when the old classifier has been discarded due to storage limitations. These synthesized pseudo-old prototypes are easily influenced by noisy samples, particularly when the intra-class variance of old classes is large. Additionally, in old prototype-based methods, the old prototypes are typically placed alongside the new classifier and directly connected to the backbone of the new model. This not only constrains the new model with the inferior prototypes of the old model, thereby limiting its capacity, but also increases the number of model parameters, posing challenges for training on large-scale datasets.

In addition to old prototype-based methods, backward-compatible training can also be achieved by considering feature distribution information at the instance level. We have found that utilizing L2 regression, coupled with appropriate constraint weights, to reconcile old and new features at the instance level results in good backward-compatible performance. However, the substantial constraints inherent in L2 regression may overly restrict the capabilities of the new model. Furthermore, in NCCL \cite{wu2022NCCL}, the authors employ contrastive learning based on a memory bank, which explicitly constrains the feature embeddings and logits of the new model by comparing them with the old ones. While NCCL provides instance-level constraints, it is limited by the size of the memory bank and the gap between old and new features; a significant gap between old and new features may limit the effectiveness of naive contrastive learning.

In an effort to overcome the limitations of previous methods, we present MixBCT, a lightweight and straightforward general framework for backward-compatible training. Our approach offers several distinct advantages: 1) It allows the new model to comprehensively obtain the distribution knowledge of the old model at the instance-level and adaptively adjust the constraints for the new model; 2) We enforce the compatibility constraint on the new classifier rather than on the new features to avoid directly impacting the backbone of the new model as much as possible; 3) Backward-compatible training can be performed using a single classification loss, and no additional parameters are necessary for the training process.

Overall, our contributions can be summarized as follows: 1) Previous methods overlooked the impact of the old model's quality on BCT tasks. We have discussed, analyzed, and validated this situation. 2) As the quality of the old model decreases, the intra-class variance of the feature increases, and previous methods struggle to perform well. MixBCT provides a solution to this problem. 3) Our approach is a lightweight and simple unified framework that requires only a single loss for backward-compatible training. This simplicity makes our method more accessible and easier to implement for practical use. 4) We extensively evaluated our proposed method on the large-scale face recognition datasets MS1Mv3 \cite{deng2019arcface} and IJB-C \cite{maze2018iarpa}, demonstrating its clear advantages.

\section{Related Works}
\label{section2} 

The backward-compatible training shares similarities with domain adaptation transfer learning\cite{you2019universal,panareda2017open,wu2024domain,ganin2015unsupervised,zhuang2020comprehensive}, incremental learning\cite{saha2020gradient,lopez2017gradient,wang2024comprehensive,zhou2023deep,castro2018end,wu2019large}, as they all require consideration of feature relationships among data. The main difference is the primary objective of backward-compatible training focuses on achieving compatibility between new and old models without relying on network structures or model initialization. In this section, we present some of the most relevant works.

\textbf{Backward-Compatible Training.} Backward-compatible training is a way to achieve the purpose of compatibility between the new model and the old model without 'backfilling'. BCT \cite{shen2020BCT} used  the classifier of the old model as a regular term to constrain the new feature. Bai et al. \cite{bai2022dual} aligned two types of embedding features by letting the class prototypes of the old and new models supervise each other's features. UniBCT \cite{zhang2022uniBCT} utilized the stronger model capability of the new model to update the old class prototypes in the learning process. AdvBCT \cite{pan2023boundary} introduces an elastic boundary constraint to further refine the alignment between the new embeddings and the old prototypes. Although using old prototypes as a regularizer can achieve backward compatibility, it suffers from a significant drawback. Specifically, it fails to capture distribution information within a category, leading to relatively large deviations in distribution between the new and old models. Such deviation is fatal in the retrieval task. NCCL \cite{wu2022NCCL} used contrastive learning. It stores the feature embeddings and logits of the old model in the memory bank, and treats the same label as positive pairs and different label as negative pairs between the old and new features. Moreover, in order to reduce the impact of new-to-old compatibility on the new feature discriminativeness, some works \cite{su2022BiCT,zhang2022darwinian,seo2023online} relaxed  the condition of ‘backfill-free’, enabling the update of old features during training.

\textbf{Cross Model Compatible.} The purpose of Cross Model Compatible (CMC) is also to achieve feature compatibility between the two different models. Unlike backward-compatible training, CMC doesn't have the concept of new and old models because the parameters of both models are fixed and cannot be learned. CMC typically achieves model compatibility by adding a feature transformation module between the two models. For example, RBT \cite{wang2020RBT} proposed a lightweight conversion module called ‘RBT’ and uses L2 similarity between features and KL divergence between logits to transform features. BC-Aligner \cite{hu2022learning} also introduces a lightweight backward compatibility transformation to align new and old embeddings. Moreover, R3AN \cite{chen2019R3AN} introduced a module composed of generative adversarial networks to align feature distributions. LCE \cite{meng2021LCE} achieves CMC through aligning class centers while restricting more compact intra-class distributions. 

\textbf{Knowledge Distillation.} Unlike model evolution in backward-compatible training, Knowledge distillation(KD) transfers the knowledge of a large-scale teacher model to a small student model. More narrowly, KD can be viewed as a form of model compression technology\cite{zhu2023survey,ma2023llm,he2023structured,shang2023post,rokh2023comprehensive}. According to the knowledge type of distillation, knowledge distillation can be categorized into response(logit)-based \cite{hinton2015KD,meng2019conditional,zhang2019fast,chen2017learning}, feature-based \cite{chen2021cross,wang2020exclusivity,wang2020exclusivity}, and relation-based \cite{peng2019CCDR,passalis2020probabilistic,chen2020learning}, etc. For example, \cite{hinton2015KD} utilized the logit of the classifier for distillation. DKD \cite{zhao2022decoupled} decoupled logit-based KD into two parts: target class knowledge distillation (TCKD) and non-target class knowledge distillation (NCKD), promising results is achieved by balancing TCKD and NCKD. Fitnets \cite{romero2014fitnets} introduced the features of the teacher hidden layer as the knowledge source for the student model. CCDR \cite{peng2019CCDR} not only transfer the instance-level information but also the correlation between instances. In general. The goal of KD is that students fully learned the knowledge of the teacher, and backward-compatible training needs to ensure that the performance of the new model is higher than the performance of the old model, whether in cross testing between the new model and the old model or in self testing.

\section{Overview}

\subsection{Problem Settings}

In general, a feature embedding model can be divided into two modules: the backbone ${\phi}$ and the classifier $\mathcal{\psi}$, the backbone ${\phi}$ maps the input $\mathcal{X}$ to a $d$ dimensional feature space $\mathcal{X} \rightarrow \mathcal{F}$, $\mathcal{F} \in \mathcal{R}^d$, while the classifier $\mathcal{\psi}$ classifies the features. In backward-compatible training, we want directly compare the features ${\phi}_{o}(x)$ generated by the old model with the features ${\phi}_{n}(x)$ generated by the new model. Following \cite{shen2020BCT}, the empirical compatibility criterion is defined as, 
\begin{equation}
   \centering
   M({\phi}_{n}(x); {\phi}_{o}(x); \mathcal{Q}; \mathcal{G}) > M({\phi}_{o}(x); {\phi}_{o}(x); \mathcal{Q}; \mathcal{G}),
   \label{metric}
\end{equation}
\noindent
where $M$ is an evaluation metric, $\mathcal{Q}$ denotes the query set, $\mathcal{G}$ denotes the gallery set. To accommodate various types of backward-compatible scenarios in practical applications(Close-Set and Open-Set), and considering that it is common for old classifier to be discarded due to storage overhead. We assume that ${\phi}_o$ of the old model is available and the old classifier $\psi_o$ is discarded. The knowledge for backward-compatible training can be extracted by the old model:  ${\phi}_{o}(\mathcal{D}_n)$, where $\mathcal{D}_n$ is the training data of the new model. 

\subsection{The Ideal Goal of Backward-Compatible Training}
\label{section_ideal_goal} 

The ultimate objective of backward-compatible training can be summarized as ensuring that any query sample $\forall\phi_n(x)$ generated by the new model satisfies the following criterion:
\begin{equation}
\begin{aligned}
   d(\phi_n(x); \phi_{o/n}(x)^P) < d(\phi_n(x); \phi_{o/n}(x)^N),   \label{Condition_total}
\end{aligned}
\end{equation}
\noindent 
where $\phi_{o/n}(x)^P$ and $\phi_{o/n}(x)^N$ are both feature embedding samples in the gallery set, $d()$ represents distance measurement function. $P$ denotes positive samples, which belong to the same category as $\phi_n(x)$, while $N$ denotes negative samples, which belong to a different category than $\phi_n(x)$. Specifically, Eq. \ref{Condition_total} can be divided into the following four constraints:
\begin{numcases}{}
   d(\phi_n(x); \phi_n(x)^P) < d(\phi_n(x); \phi_n(x)^N) \label{Condition_1} \\ 
   d(\phi_n(x); \phi_n(x)^P) < d(\phi_n(x); \phi_o(x)^N) \label{Condition_2} \\ 
   d(\phi_n(x); \phi_o(x)^P) < d(\phi_n(x); \phi_n(x)^N) \label{Condition_3} \\   
   d(\phi_n(x); \phi_o(x)^P) < d(\phi_n(x); \phi_o(x)^N) \label{Condition_4}. 
\end{numcases}

\subsection{Analysis of Old Prototype-based and Instance-based Methods}

Based on four constraints above, we will conduct a brief analysis of current old prototype-based and instance-based backward-compatible training methods. We assume that the new model can be ideally trained, i.e., the constraint of Eq. \ref{Condition_1} can be met, and we only analyze backward compatibility constraints Eq. \ref{Condition_2}, \ref{Condition_3} and \ref{Condition_4} which involving the interaction of old and new features.

The old prototype-based methods \cite{bai2022dual,shen2020BCT,zhang2022uniBCT,pan2023boundary} utilize the class center of the old feature as the backward-compatible constraint item and included the following constraint:
\begin{equation}
\begin{aligned}
   d(\phi_n(x); \ \mathcal{C}_o^P) < d(\phi_n(x); \  \mathcal{C}_o^N),
   \label{BCT_c}
\end{aligned}
\end{equation}
where $\mathcal{C}_o^P$ and $\mathcal{C}_o^N$ denotes the positive and negative class prototypes of the old features, respectively. Obviously, the new model can only access the old class prototypes as a source of information, and the Eq. \ref{BCT_c} can not guarantee constraints Eq. \ref{Condition_2}, \ref{Condition_3} and \ref{Condition_4}. 

In light of the issues with ignoring old feature distribution information in old prototype-based methods, instance-based methods such as L2 regression appear to be a viable option, instance-level alignment can effectively fit the feature distribution of the old model. The backward-compatible loss based on L2 regression can be formulated as,
\begin{equation}
    %\begin{aligned}
     \mathcal{L}_{L_2BCT} = \mathcal{L}_{cls} + \lambda \mathcal{L}_{2}, \quad where \ \mathcal{L}_{2} =  \frac{1}{N} \sum_{i=1}^{N} \left \| \phi_{n}(x_i), \phi_{o}(x_i) \right \|_2,
    %\end{aligned}
    \label{L2}
\end{equation}
where $\mathcal{L}_{cls}$ is the classification loss, $\left \| \bullet \right \|_2$ denote L2 distance, and $N$ indicates the batch size.  the training process of L2 regression is simple and does not require additional parameters. However, the constraint imposed by $\lambda\mathcal{L}_2$ does also not align with the constraints of Eq.  \ref{Condition_2}, \ref{Condition_3}, and \ref{Condition_4}.

\subsection{Mixing Old and New Features for Backward-Compatible Training}

The old prototype-based approach is unsuitable for dealing with the relatively large intra-class variance of the old features. Although instance-level methods such as L2 regression can capture the distribution information of the features of the previous model, its strong constraint limits the efficacy. To address these issues, we propose a simple yet effective approach that strikes a balance between the methods of old prototype-based and instance-based while also going one step further. We call this approach ‘MixBCT’.

\begin{figure}[htbp]
\centering
\setlength{\abovecaptionskip}{1pt}
\includegraphics[width=0.6\textwidth]{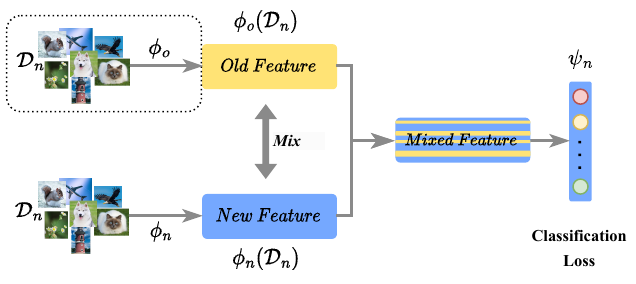}
\caption{During the training phase, the old features are extracted by $\phi_o(\mathcal{D}_n)$ and mix them with new features $\phi_n(\mathcal{D}_n)$ to be fed into the new classifier $\psi_n$. Once backward-compatible training is complete, only the new model is required for feature extraction. It's worth noting that the dashed part of our workflow can be completed prior to conducting backward-compatible training.}
\label{our_framework}
\end{figure}

Fig. \ref{our_framework} illustrates the workflow of MixBCT. During each step of the training phase, we randomly mix the new and old features in a proportional manner and classify them using the new model's classifier. The $Mix()$ operation can be represented as:
\begin{equation}
\begin{aligned}
    \phi(x)^{Mix} = Mix(\alpha \{\phi_o(x)\}, \  (1-\alpha)\{\phi_n(x)\} ; \ x \in \mathcal{D}_n), 
\end{aligned}
\end{equation}
\noindent 
where $\alpha$ is a ratio factor, and the $Mix()$ operation mix old features and new features in proportion by replacing some of the new features with the old ones. For example, suppose the batch-size is $B$, in training  process, we will replace $\lfloor\alpha \times B\rfloor$ of new features with the old features. The optimization objective is:
\begin{equation}
\begin{aligned}
     w^{\phi}_n, \ w^{\psi}_n = arg \ min\mathcal{L}_{MixBCT}( \ \psi({\phi(x)^{Mix}})),
     \label{Optimization_obj}
\end{aligned}
\end{equation}

where $w_n$ denotes the parameters of the new model. Eq. \ref{Optimization_obj} means that we can achieve backward compatibility with a single classification loss function:
\begin{equation}
\mathcal{L}_{MixBCT} = L_{cls}(\phi(x)^{Mix}).
   \label{Loss_mixBCT_c}
\end{equation}

\begin{figure*}[htbp]
    \centering 
    \captionsetup[subfigure]{labelformat=empty}
    \setlength{\abovecaptionskip}{-10pt}
    \subfloat[]{\includegraphics[width=0.195\textwidth]{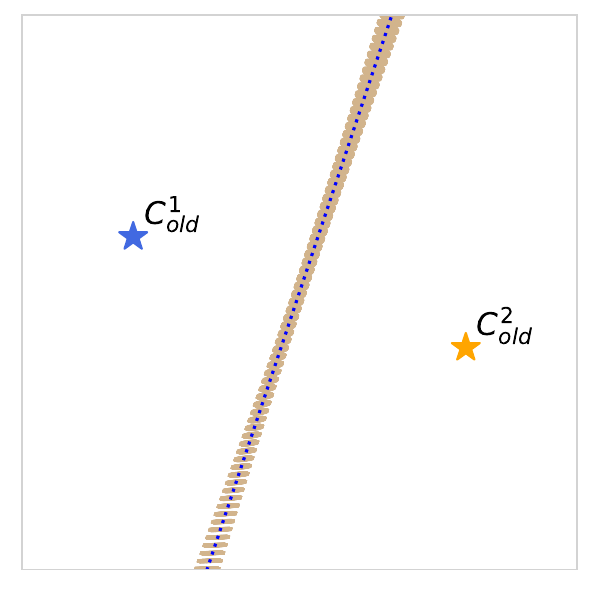}}
    \label{fig1}
    \subfloat[]{\includegraphics
         [width=0.195\textwidth]{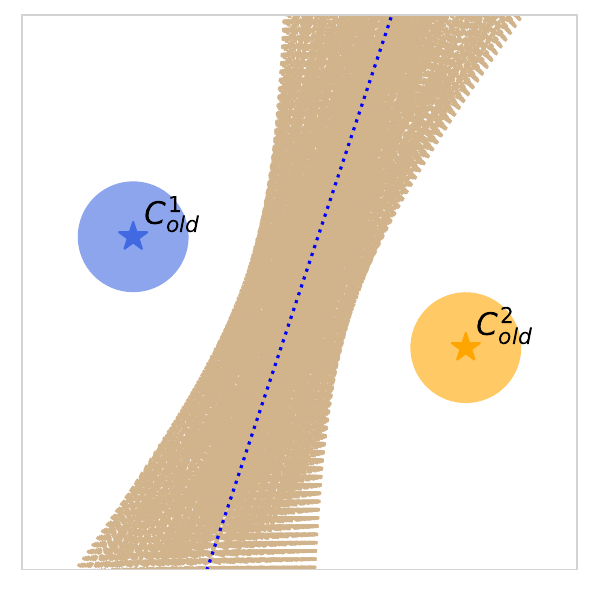}}
         \label{fig2}
    \subfloat[]{\includegraphics
         [width=0.195\textwidth]{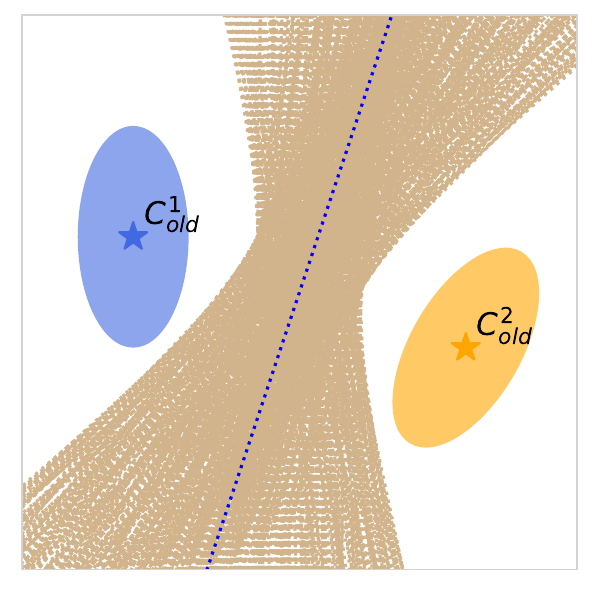}}
         \label{fig3}
    \subfloat[]{\includegraphics
         [width=0.195\textwidth]{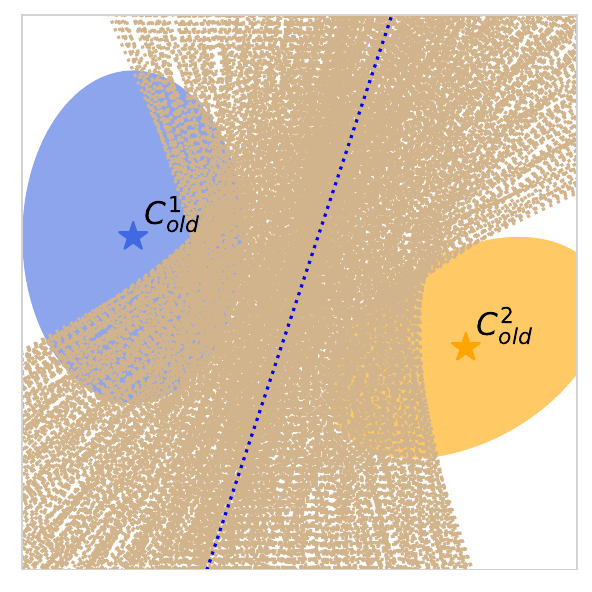}}
         \label{fig4}
    \subfloat[]{\includegraphics
         [width=0.195\textwidth]{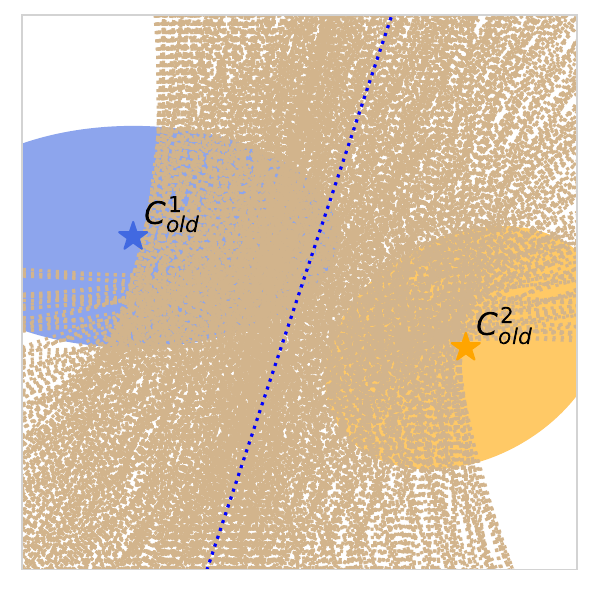}}
         \label{fig5}
    \caption{Generally, consider any two classes $C^1$ and $C^2$, in each subgraph, the circle or ellipse is the distribution areas of the old features, $C_{old}^1$ and $C_{old}^2 $ are the old class centers of class $C^1$ and class $C^2$ respectively. As we move from left to right, the quality of the old model decreases progressively. The light blue dashed line is the boundary hyperplane used by the current old prototype-based method to constrain the new feature, and it is the midperpendicular of the line between two classes. Specifically, the old prototype-based method restricts new features belong to $C^1$ to the left of the light blue dashed line, and new features belong to $C^2$ to the right. The light orange dashed line represents the boundary hyperplane where our method constrains the new model's classifier weights. The MixBCT imposes the constraints: $d(\phi_{o/n}^1(x); C_{new}^1) < d(\phi_{o/n}^1(x); C_{new}^2)$ and $d(\phi_{o/n}^2(x); C_{new}^2) < d(\phi_{o/n}^2(x); C_{new}^1)$, it contains constraints: $d(\phi_{o}^1(x); C_{new}^1) < d(\phi_{o}^1(x); C_{new}^2)$ and $d(\phi_{o}^2(x); C_{new}^2) < d(\phi_{o}^2(x); C_{new}^1)$. Thus, the area not enclosed by the light orange dashed line forms the feasible domain.}
    \label{fig:constraints}
\end{figure*}

\textbf{Analysis}: MixBCT introduces constraints on the classifier of the new model to circumventing the direct limitations imposed by the old prototype-based framework on the new model's backbone. Fig. \ref{fig:constraints} visually illustrates the constraints imposed by MixBCT and the old prototype-based approach. We can see that MixBCT considers the distribution of the old features and dynamically adjusts the constraint domain for the new model. In cases where the old feature class is compact, MixBCT places fewer constraints on the new model to maximize its potential while ensuring backward compatibility. Conversely, as the old features within the class become more dispersed, MixBCT imposes stronger constraints on the new model to ensure good performance for backward retrieval. In contrast, the old prototype-based method enforces the same constraints on the new model throughout, thus making it challenging to guarantee backward-compatible performance when the old model has lower quality.

Taking an alternative perspective, let's assume that the new model is optimally trained and adequately robust. In MixBCT, it employs a single classification loss function that incorporates both old and new features for embedding representative learning in retrieval tasks. The primary objective of this classification loss is to ensure that, for the mixed features, the maximum intra-class distance is always less than the minimum inter-class distance: $d(\phi_{o/n}(x); \phi_{o/n}(x)^P) < d(\phi_{o/n}(x); \phi_{o/n}(x)^N)$, where the distance relationship between the old features is fixed. It is obvious that the optimization objective of the classification loss function Eq. \ref{Loss_mixBCT_c} encompasses the ideal goal of backward-compatible training described in  Eq. \ref{Condition_total}: $d(\phi_n(x); \phi_{o/n}(x)^P) < d(\phi_n(x); \phi_{o/n}(x)^N)$. Consequently, MixBCT can effectively achieve the ideal objective of backward-compatible training.

Additionally, to mitigate the impact of noise in the old feature $\phi_o(\mathcal{D}_n)$ caused by the low quality of the old model, similar to UniBCT \cite{zhang2022uniBCT}, we utilize normalized Euclidean distance to identify noisy samples. Specifically, L2 normalization is performed on $\phi_o(\mathcal{D}_n)$ along each dimension to standardize its scale. Next, we compute the Euclidean distance of each sample to the center of its corresponding class and exclude the top 10\% of the old features that are farthest from the center. With the inclusion of the denoised old feature $\phi_o(x)^*$ into the mixed features $\phi(x)^{Mix^*}$, the resulting loss function after denoising the old feature is:
\begin{equation}
\mathcal{L}_{MixBCT} = L_{cls}(\phi(x)^{Mix^*}).
   \label{Loss_mixBCT_c_Denois}
\end{equation}

The simple workflow of the MixBCT makes it can be easily foreseen that our method basically not affect the training speed of the new model during the training process and does not require additional parameter. This simplicity makes our method more accessible and easier to implement for practical use.

\section{Experiments}

\subsection{Datasets and Evaluation}

\textbf{Datasets: }Large-scale datasets are necessary for evaluating the performance of backward-compatible training since one of its goal is to avoid updating old features in the gallery set. We employ the MS1Mv3 \cite{deng2019arcface} dataset containing 5,179,510 face images and 93,431 unique identities as the training set to train the feature embedding model. We evaluate the backward-compatible performance using the widely used and challenging IJB-C \cite{maze2018iarpa} face recognition benchmark. To assess the performance, we conduct both self-test and cross-test, with the former examining the impact of backward compatibility on the performance of new model and the latter assessing the backward-compatible performance.

\noindent  
\textbf{Metric: }The IJB-C 1:1 verification protocol comprises a large set of 469,376 template pairs, where the task is to determine whether a given pair of templates corresponds to the same identity. In cross-test, we generate the first template using the old model and the second template using the new model. We evaluate performance using the TAR@FAR(TAR: True Acceptance Rate, FAR: False Acceptance  Rate) metric for the 1:1 verification protocol. In the IJB-C 1:N identification protocol, the gallery set contains 3,531 templates, while the query set comprises 19,593 templates. The task is to use the templates in the query set to retrieve the corresponding templates in the gallery set. In cross-test of 1:N protocol, we use the old model to generate the gallery set and the new model to generate the query set. Note that the IJB-C benchmark is an open-set evaluation dataset, where some of the identities in the query set may not exist in the gallery set. We use the TPIR@FAR(TPIR: True Positive Identification Rate) as the evaluation metric under the open-set setting for the 1:N protocol.

\noindent  
\textbf{Lower Bound and Upper Bound: } We trained the old model using iResnet18 \cite{duta2021improved} and the new model using iResnet50 \cite{duta2021improved}. The performance of the old model trained on $\mathcal{D}_o$ will serve as the lower bound. If the cross-test performance exceeds the lower bound, we can consider that the goal of achieving backward-compatible has been met. Previous works used the self-test performance of the new model trained on $\mathcal{D}_n$ without the constraint as the upper bound. However, since the information of the old model is used in the backward-compatible training process of the new model, unlike previous works, we use the self-test performance of the new model trained on $\mathcal{D}^{100\%} = \mathcal{D}_o \cup \mathcal{D}_n$ rather than just $\mathcal{D}_n$ as the upper bound.

\subsection{Benchmarks}
\label{Benchmarks}
Our method is suitable for both close-set and open-set model upgrading scenarios. We consider four specific model upgrade scenarios with varying old model qualities. The upgrade scenarios include: 1)\textbf{Extended-Data}: The top 30\% data of each ID is used to train the old model, while the whole dataset is used to train the new model. 2) \textbf{Extended-Class}: The top 30\% IDs are used to train the old model, while the whole dataset is used to train the new model. 3) \textbf{Open-Data}: The top 30\% data of each ID  is used to train the old model, while the remaining 70\% data is used to train the new model. 4) \textbf{Open-Class}: The top 30\% IDs are used to train the old model, while the remaining 70\% IDs are used to train the new model. The details of these dataset settings are provided in Tab. \ref{data_split}.

\begin{table}[htbp]
\caption{Details of each training set split. }
\centering
%\renewcommand{\arraystretch}{0.9}
%\resizebox{\linewidth}{!}{
\begin{tabular}{lc||ccc}
\toprule
 Setup & Training-set  &  Subset & \#images & \#classes \\
 \midrule
Extended-Data & $\mathcal{D}_o^{30\%d}$ : $\mathcal{D}^{100\%}$  & $\mathcal{D}_o^{30\%d}$ & 1,554,138 & 93,431 \\
Extended-Class & $\mathcal{D}_o^{30\%c}$ : $\mathcal{D}^{100\%}$  &$\mathcal{D}_n^{70\%d}$ & 3,625,384 & 93,431 \\
Open-Data & $\mathcal{D}_o^{30\%d}$ : $\mathcal{D}_n^{70\%d}$  &$\mathcal{D}_o^{30\%c}$  & 1,581,241 & 28,029 \\
Open-Class & $\mathcal{D}_o^{30\%c}$ : $\mathcal{D}_n^{70\%c}$  &$\mathcal{D}_n^{70\%c}$ & 3,598,269 & 65,402 \\
%$I$-$D$ & $\mathcal{D}_o^{30\%d}$ : $\mathcal{D}_n^{30\%d}$  %&$\mathcal{D}^{100\%}$ & 5,179,510 & 93,431 \\
\bottomrule
\end{tabular}
%}

\label{data_split}
\end{table}

%appendix materials.

\subsection{Implementation Details}
We use 8 NVIDIA 2080Ti/3090Ti GPUs for training and apply automatic mixed precision (AMP) with float16 and float32. We use standard stochastic gradient descent (SGD) as the optimizer. Batch-size is set to 128 $\times$ 8. An initial learning rate of 0.1, and the learning rate linearly decays to zero over the course of training. The weight decay is set to $5\times10^{-4}$ and momentum is 0.9. The training stops after 35 epochs. We set the ratio $\alpha$ of old and new features in the $Mix()$ process to 0.3. Moreover, the $\lambda$ in L2 regression is set to 10 to match loss scale with the classfication loss after extensive testing.

\subsection{Performance Comparison in Various Old Model Qualities}

The selection of loss function and embedding dimension greatly impacts the quality of the model. Therefore, we trained the old model using popular embedding loss Arcface \cite{deng2019arcface} and the simple Cross-Entropy loss, setting feature dimensions with 128, 256, or 512 to construct models of varying qualities. A lower performance of the old model indicates inferior quality in terms of feature embeddings. For the new model, we trained using Arcface loss with the same feature dimensions as the old model. We first evaluate the performance under the open-class scenario, which is considered the most challenging scenario of all.

\begin{wraptable}{r}{7.3cm}
\vspace{-1.35em}
\caption{Comparison of our method and state-of-the-art methods in the \textbf{Open-Class scenario with different quality} of old model. From top to bottom, the quality of the old model gradually improved. }
\begin{center}
\resizebox{\linewidth}{!}{
\begin{tabular}{lccccc}
\hline
\multicolumn{1}{l|}{}  & \multicolumn{2}{c|}{1:1 Verification} & \multicolumn{2}{c|}{1:N Identification}  \\ \cline{2-5}
\multicolumn{1}{l|}{}  & \multicolumn{1}{c|}{CT} & \multicolumn{1}{c|}{ST} & \multicolumn{1}{c|}{CT} & \multicolumn{1}{c|}{ST}  \\ \cline{2-5}
\multicolumn{1}{l|}{\multirow{-3}{*}{Method}} & \multicolumn{2}{c|}{TAR@FAR=10-4} & \multicolumn{2}{c|}{TPIR@FAR=10-2}  & \multicolumn{1}{c}{\multirow{-3}{*}{AVG}} \\ \hline
\rowcolor[gray]{0.9}
\multicolumn{6}{c}{Open-Class  \qquad \qquad  128-Dim    \qquad \qquad  $\mathcal{L}_{CE}$ / $\mathcal{L}_{Arc}$}   \\ 
\hline
$Lower$-$B$ & {-} & {\color[HTML]{FF0000} 0.6358} & {-} & {\color[HTML]{FF0000} 0.4280} & {-} \\
$Upper$-$B^{\prime}$  & 0 & 0.9528 & 0 & 0.9269 & - \\  
$Upper$-$B$  & {0} & {\color[HTML]{4472C4} 0.9622} & {0} & {\color[HTML]{4472C4} 0.9419} & {\color[HTML]{4472C4} -}  \\
$BCT$ & 0.6263 & 0.9553 & 0.4251 & 0.9288 & 0.7339 \\
$UniBCT$  & 0.7430 & 0.9483 & 0.5839 & 0.9190  & 0.7986 \\
$NCCL$  & 0.7565 & 0.9446 & 0.5702 & 0.9153   & 0.7967 \\
$AdvBCT$ &  0.7515 & 0.9456 & 0.5933 & 0.9149 &  0.8013 \\
$L_2$  & 0.8094 & 0.9362 & 0.6747 & 0.8959  & 0.8291 \\
$MixBCT$  & 0.8305 & 0.9525 & 0.6973 & 0.9224 & \textbf{0.8507} \\ \hline

\rowcolor[gray]{0.9}
\multicolumn{6}{c}{Open-Class  \qquad \qquad  256-Dim   \qquad \qquad   $\mathcal{L}_{CE}$ / $\mathcal{L}_{Arc}$} \\ \hline
$Lower$-$B$ & {-} & {\color[HTML]{FF0000} 0.8236} & {-} & {\color[HTML]{FF0000} 0.6929} & {-}\\
$Upper$-$B$   & {0} & {\color[HTML]{4472C4} 0.9649} & {0} & {\color[HTML]{4472C4} 0.9447} & {\color[HTML]{4472C4} -}  \\
$BCT$ & 0.8474 & 0.9568 & 0.7407 & 0.9321 & 0.8693\\
$UniBCT$ & 0.9003 & 0.9517 & 0.8271 & 0.9238  & 0.9007\\
$NCCL$  & 0.8970 & 0.9493 & 0.8262 & 0.9190  & 0.8979 \\
$AdvBCT$ &  0.8824 & 0.9465 & 0.8003 & 0.9148 &  0.8860 \\
%$AdvBCT5$ &  0.8792 & 0.9415 & 0.7800 & 0.9001 &  0.8752 \\
$L_2$  & 0.8967 & 0.9422 & 0.8235 & 0.9055 & 0.8920 \\
$MixBCT$  & 0.9094 & 0.9510 & 0.8474 & 0.9256  & \textbf{0.9084} \\ \hline
\rowcolor[gray]{0.9}
\multicolumn{6}{c}{Open-Class \qquad \qquad 512-Dim   \qquad \qquad $\mathcal{L}_{Arc}$ / $\mathcal{L}_{Arc}$} \\ \cline{1-6}
$Lower$-$B$  & {-} & {\color[HTML]{FF0000} 0.9250} & {-} & {\color[HTML]{FF0000} 0.8783} & {-}\\
$Upper$-$B$  & {0.0002} & {\color[HTML]{4472C4} 0.9647} & {0} & {\color[HTML]{4472C4} 0.9453} & {-}\\
$BCT$ & 0.9320 & 0.9582 & 0.8937 & 0.9366 &  0.9301 \\
$UniBCT$ & 0.9395 & 0.9596 & 0.9057 & 0.9372 &  0.9355 \\
$NCCL$ &  0.9405 & 0.9587 & 0.9058 & 0.9344 &  0.9349 \\
$AdvBCT$ &  0.9357 & 0.9563 & 0.8976 & 0.9288 &  0.9296 \\
%$AdvBCT5$ &  0.9352 & 0.9560 & 0.8936 & 0.9294 &  0.9286 \\
$L_2$ &  0.9392 & 0.9574 & 0.9032 & 0.9336 &  0.9334 \\
$MixBCT$ & 0.9410 & 0.9597 & 0.9071 & 0.9367  &  \textbf{0.9361}\\ \hline
\end{tabular}
}

\label{Tabel_OC}
\end{center}
\end{wraptable}

The experimental results are presented in Tab. \ref{Tabel_OC}. Here, ‘CT' and ‘ST' refers to the cross-test and self-test, respectively. ‘Open-Class' denotes the Open-Class scenario. ‘$\mathcal{L}_{CE}$ / $\mathcal{L}_{Arc}$' indicates that the old model uses Cross-Entropy loss while the new model uses Arcface loss. ‘128-dim' denotes the dimension of the feature embedding as 128. ‘$Lower$-$B$' represents the lower bound. ‘$Upper$-$B^{\prime}$' signifies the upper bound utilized by the previous methods. ‘$Upper$-$B$' denotes the real upper bound. ‘AVG' stands for the average of the performance, which is a simple statistic, and ‘$L_2$' refers to L2 regression.

We compared the state-of-the-art methods\cite{wu2022NCCL,pan2023boundary,zhang2022uniBCT,shen2020BCT}, including L2 regression. In particular, when the old model employs Cross-Entropy loss and feature dimension is set to 128, the quality of the old model is relatively low. Our proposed method achieves significant improvement compared with state-of-the-art methods. UniBCT \cite{zhang2022uniBCT} mitigated the problem of inaccurate old class prototype in BCT \cite{shen2020BCT} to some extent by updating old feature class prototypes with the new model, while AdvBCT \cite{pan2023boundary} adding further constraints. However, it's still limited to the old prototype during backward-compatible training. NCCL \cite{wu2022NCCL}, in all experiments, we increased its memory bank size to 8192 instead of 2048 in order to make it work better with large-scale datasets, utilized both old and new features and logit for contrastive learning.  Nevertheless, due to the difference in capability between the old and new models, the gap between the feature space they learn is relatively large, making it challenging to optimize instance-level contrast learning directly. 
In the 1:1 verification and 1:N identification task of cross-test, MixBCT achieves 7.4\% and 10.4\% higher than the previous best performing methods, respectively. In self-test both 1:1 verification task and 1:N identification task, consistent performance improvements were also achieved. 

It can be observed that L2 regression, coupled with appropriate constraint weights, achieves relatively better cross-test performance compared to previous methods when the old model is of relatively low quality. However, its strong constraint can negatively impact the performance of the new model in self-test. 
%because of it directly aligns the old and new feature distributions

Moreover, as the quality of the old model improves, the performance gap between the various methods is reduced, including L2 regression. This is  due to the old features were already well-learned and the gap between the old and new models was smaller. In such scenarios, the optimization objectives of different methods converged towards a similar goal, i.e., aligning the class centers between the old and new models.

\subsection{Performance Comparison in Various Backward-Compatible Scenarios}
Our proposed method has broad applicability and can be utilized in various scenarios, In addition to the Open-class scenario, we have compared the performance of our method with the state-of-the-art methods in other scenarios, including Extended-Class, Open-Data, and Extended-Data, which basically cover all the scenarios that require backward-compatible training. Details of these scenarios are described in section \ref{Benchmarks}.

\begin{table}[htbp]
\caption{Performance comparison with state-of-the-art methods in various backward-compatible training scenarios and the old models have relatively \textbf{Low-Quality} and \textbf{High-Quality} .} %‘Extended-Class’: Extended-Class scenario. ‘Open-Data’: Open-Data scenario. ‘Extended-Data’: Extended-Data scenario.}
\begin{center}
\resizebox{\linewidth}{!}{
\begin{tabular}{l|ccccc|ccccc}
\hline
\multicolumn{1}{l|}{}  & \multicolumn{2}{c|}{1:1 Verification} & \multicolumn{2}{c|}{1:N Identification} &  \multicolumn{1}{l|}{}  & \multicolumn{2}{c|}{1:1 Verification} & \multicolumn{2}{c|}{1:N Identification}  \\ \cline{2-5}  \cline{7-10}
\multicolumn{1}{l|}{}  & \multicolumn{1}{c|}{CT} & \multicolumn{1}{c|}{ST} & \multicolumn{1}{c|}{CT} & \multicolumn{1}{c|}{ST}  & \multicolumn{1}{l|}{}  & \multicolumn{1}{c|}{CT} & \multicolumn{1}{c|}{ST} & \multicolumn{1}{c|}{CT} & \multicolumn{1}{c|}{ST}  \\ \cline{2-5} \cline{7-10} 
\multicolumn{1}{l|}{\multirow{-3}{*}{Method}} & \multicolumn{2}{c|}{TAR@FAR=10-4} & \multicolumn{2}{c|}{TPIR@FAR=10-2}  & \multicolumn{1}{c}{\multirow{-3}{*}{AVG}} & \multicolumn{2}{|c|}{TAR@FAR=10-4} & \multicolumn{2}{c|}{TPIR@FAR=10-2}  & \multicolumn{1}{c}{\multirow{-3}{*}{AVG}} \\ \hline

\multicolumn{1}{c|}{}  & \multicolumn{5}{c|}{Low-Quality}  & \multicolumn{5}{c}{High-Quality}  \\ \hline

\rowcolor[gray]{0.9}
\multicolumn{1}{c}{}  & \multicolumn{5}{c|}{Extended-Class \qquad  128-Dim  \qquad $\mathcal{L}_{CE}$ / $\mathcal{L}_{Arc}$}  & \multicolumn{5}{c}{Extended-Class \qquad  512-Dim   \qquad $\mathcal{L}_{Arc}$ / $\mathcal{L}_{Arc}$} \\ \hline
$Lower$-$B$ & {-} & {\color[HTML]{FF0000} 0.6358} & {-} & {\color[HTML]{FF0000} 0.4280} & {-} & {-} & {\color[HTML]{FF0000} 0.9250} & {-} & {\color[HTML]{FF0000} 0.8783}  & {-}\\
$Upper$-$B$ &  {0} & {\color[HTML]{4472C4} 0.9622} & {0} & {\color[HTML]{4472C4} 0.9419}  & {\color[HTML]{4472C4} -}& {0.0002} & {\color[HTML]{4472C4} 0.9647} & {0} & {\color[HTML]{4472C4} 0.9453} & {\color[HTML]{4472C4} -}\\
$BCT$ &  0.6400 & 0.9605 & 0.4474 & 0.9402 & 0.7470 &  0.9350 & 0.9637 & 0.8962 & 0.9442 & 0.9348 \\
$UniBCT$   & 0.7595 & 0.9542 & 0.6033 & 0.9289 & 0.8115  & 0.9429 & 0.9637 & 0.9115 & 0.9440 & 0.9405\\
$NCCL$  & 0.7624 & 0.9536 & 0.5784 & 0.9284 & 0.8057 & 0.9430 & 0.9624 & 0.9120 & 0.9432 & 0.9402 \\
$AdvBCT$ &  0.7628 & 0.9550 & 0.6196 & 0.9307 &  0.8170 &  0.9411 & 0.9615 & 0.9085 & 0.9401 &  0.9378 \\
$L_2$  & 0.8228 & 0.9497 & 0.6907 & 0.9207 & 0.8460 & 0.9426 & 0.9601 & 0.9105 & 0.9397 & 0.9382 \\
$MixBCT$  & 0.8363 & 0.9580 & 0.7104 & 0.9329 & \textbf{0.8594} & 0.9450 & 0.9637 & 0.9122 & 0.9441  & \textbf{0.9413} \\ \hline

\rowcolor[gray]{0.9}
\multicolumn{1}{c}{} & \multicolumn{5}{c|}{Open-Data \qquad  512-Dim  \qquad  $\mathcal{L}_{CE}$ / $\mathcal{L}_{Arc}$} & \multicolumn{5}{c}{Open-Data \qquad   512-Dim  \qquad  $\mathcal{L}_{Arc}$ / $\mathcal{L}_{Arc}$}  \\ \hline
$Lower$-$B$ & {-} & {\color[HTML]{FF0000} 0.8076} & {-} & {\color[HTML]{FF0000} 0.6670} & {-}  &  {-} & {\color[HTML]{FF0000} 0.9295} & {-} & {\color[HTML]{FF0000} 0.8911} & {-} \\
$Upper$-$B$  & {0.0002} & {\color[HTML]{4472C4} 0.9647} & {0} & {\color[HTML]{4472C4} 0.9453}  & {\color[HTML]{4472C4} -}  & {0.0002} & {\color[HTML]{4472C4} 0.9647} & {0} & {\color[HTML]{4472C4} 0.9453}  & {\color[HTML]{4472C4} -}  \\

$BCT$   & 0.8213 & 0.9619 & 0.6890 & 0.9424 & 0.8537 &  0.9382 & 0.9637 & 0.9053 & 0.9437 & 0.9377\\
$UniBCT$ & 0.8937 & 0.9592 & 0.8186 & 0.9380  & 0.9024 & 0.9452 & 0.9622 & 0.9178 & 0.9427  & 0.9420\\
$NCCL$  & 0.8929 & 0.9546 & 0.8147 & 0.9303  & 0.8981 &   0.9466 & 0.9635 & 0.9167 & 0.9440  & 0.9427\\
$AdvBCT$ &  0.8800 & 0.9556 & 0.8019 & 0.9303 &  0.8920 &  0.9444 & 0.9617 & 0.9132 & 0.9401 &  0.9399\\
$L_2$  & 0.8966 & 0.9476 & 0.8203 & 0.9171 & 0.8954 &  0.9473 & 0.9615 & 0.9183 & 0.9427  & 0.9424 \\
$MixBCT$ & 0.9044 & 0.9551 & 0.8410 & 0.9309 & \textbf{0.9079} &  0.9467 & 0.9628 & 0.9178 & 0.9453  & \textbf{0.9432} \\ \hline
\rowcolor[gray]{0.9}

\multicolumn{1}{c}{} & \multicolumn{5}{c|}{Extended-Data  \qquad  512-Dim  \qquad  $\mathcal{L}_{CE}$ / $\mathcal{L}_{Arc}$} & \multicolumn{5}{c}{Extended-Data \qquad  512-Dim  \qquad  $\mathcal{L}_{Arc}$ / $\mathcal{L}_{Arc}$} \\ \hline
$Lower$-$B$ & {-} & {\color[HTML]{FF0000} 0.8076} & {-} & {\color[HTML]{FF0000} 0.6670} & {-} & {-} & {\color[HTML]{FF0000} 0.9295} & {-} & {\color[HTML]{FF0000} 0.8911}  & {-}\\
$Upper$-$B$ & {0.0002} & {\color[HTML]{4472C4} 0.9647} & {0} & {\color[HTML]{4472C4} 0.9453}  & {\color[HTML]{4472C4} -}  & {0.0002} & {\color[HTML]{4472C4} 0.9647} & {0} & {\color[HTML]{4472C4} 0.9453}  & {\color[HTML]{4472C4} -} \\

$BCT$ &  0.8217 & 0.9631 & 0.6891 & 0.9425  & 0.8541 & 0.9379 & 0.9629 & 0.9060 & 0.9450 & 0.9380  \\
$UniBCT$ & 0.8946 & 0.9610 & 0.8160 & 0.9396 & 0.9028 & 0.9467 & 0.9637 & 0.9163 & 0.9443 & 0.9428\\
$NCCL$ & 0.8923 & 0.9548 & 0.8180 & 0.9292 & 0.8986 & 0.9467 & 0.9628 & 0.9196 & 0.9451  & 0.9436\\
$AdvBCT$ &  0.8838 & 0.9567 & 0.8039 & 0.9302 &  0.8937 &  0.9450 & 0.9611 & 0.9150 & 0.9436 &  0.9412\\
$L_2$ &  0.8965 & 0.9491 & 0.8272 & 0.9187 & 0.8979 & 0.9473 & 0.9610 & 0.9177 & 0.9440 & 0.9425 \\
$MixBCT$ & 0.9067 & 0.9570 & 0.8440 & 0.9329  & \textbf{0.9102}& 0.9476 & 0.9648 & 0.9186 & 0.9453 & \textbf{0.9441}  \\ \hline
\end{tabular}
}

\label{Table_lowq}
\end{center}
\end{table}

We conducted various scenario experiments under two conditions: first, when the old model had relatively low quality, and second, when it had relatively high quality. It should be noted that, in the Extended-Data and Open-Data scenarios, as well as in low-quality model setting, we opted to set the feature dimension of the old model to 512 instead of 128. This was necessary due to the significantly poor performance observed when training the old model with a feature dimension of 128, which makes it unlikely to be adopted in real-world retrieval systems. The experimental results are presented in Tab. \ref{Table_lowq}.

As we can see, when applied to an old model with low-quality, MixBCT maintains a clear performance advantage across various scenarios, with the advantage becoming increasingly apparent as the quality of the old model decreases. These observations align with the findings from the Open-Class scenario. When applied to a high-quality old model, our approach maintains the high performance of backward compatibility, and achieving consistent performance improvements over the state-of-the-art approachs. These results provide strong evidence of the robustness and effectiveness of our proposed method across a range of scenarios.

\subsection{Ablation Study}

\paragraph{The Effect of Hyperparameter $\alpha$.}

\begin{wraptable}{r}{7.3cm}
\vspace{-1.5em}
\caption{The effect of the hyperparameter $\alpha$ under \textbf{different quality} of old model.}
\centering
%\begin{center}
\resizebox{\linewidth}{!}{

\begin{tabular}{lccccc}
\hline
\multicolumn{1}{l|}{}  & \multicolumn{2}{c|}{1:1 Verification} & \multicolumn{2}{c|}{1:N Identification}  \\ \cline{2-5}
\multicolumn{1}{l|}{}  & \multicolumn{1}{c|}{CT} & \multicolumn{1}{c|}{ST} & \multicolumn{1}{c|}{CT} & \multicolumn{1}{c|}{ST}  \\ \cline{2-5}
\multicolumn{1}{l|}{\multirow{-3}{*}{Setup}} & \multicolumn{2}{c|}{TAR(\%)@FAR=10-4} & \multicolumn{2}{c|}{TPIR(\%)@FAR=10-2}  & \multicolumn{1}{c}{\multirow{-3}{*}{AVG}} \\ \hline

\rowcolor[gray]{0.9}
\multicolumn{6}{c}{Open-Class  \qquad \qquad  128-Dim   \qquad \qquad  $\mathcal{L}_{CE}$ / $\mathcal{L}_{Arc}$} \\ \hline
$Lower$-$B$ & {-} & {\color[HTML]{FF0000} 0.6358} & {-} & {\color[HTML]{FF0000} 0.4280} & {-} \\
 
$Upper$-$B$  & {0} & {\color[HTML]{4472C4} 0.9622} & {0} & {\color[HTML]{4472C4} 0.9419} & {\color[HTML]{4472C4} -}  \\

$\alpha=0.1$ & 0.7876 & 0.9568 & 0.6308 & 0.9308  & 0.8265 \\ 
$\alpha=0.2$ & 0.8172 & 0.9557 & 0.6730 & 0.9285  & 0.8436 \\ 
$\alpha=0.3$ & 0.8305 & 0.9525 & 0.6973 & 0.9224  & \textbf{0.8507} \\ 
$\alpha=0.4$ & 0.8311 & 0.9459 & 0.6932 & 0.9091  & 0.8448 \\ 
$\alpha=0.5$ & 0.8166 & 0.9378 & 0.6821 & 0.8953  & 0.8330 \\ 

\hline
\rowcolor[gray]{0.9}
\multicolumn{6}{c}{Open-Class  \qquad \qquad  512-Dim   \qquad \qquad   $\mathcal{L}_{Arc}$ / $\mathcal{L}_{Arc}$} \\ \hline

$Lower$-$B$  & {-} & {\color[HTML]{FF0000} 0.9250} & {-} & {\color[HTML]{FF0000} 0.8783} & {-}\\
$Upper$-$B$  & {0.0002} & {\color[HTML]{4472C4} 0.9647} & {0} & {\color[HTML]{4472C4} 0.9453} & {\color[HTML]{4472C4} -}\\
$\alpha=0.1$ & 0.9401 & 0.9594 & 0.9073 & 0.9364  & 0.9358 \\ 
$\alpha=0.2$ & 0.9406 & 0.9590 & 0.9087 & 0.9380  & \textbf{0.9366} \\ 
$\alpha=0.3$ & 0.9410 & 0.9597 & 0.9071 & 0.9367  & 0.9361 \\ 
$\alpha=0.4$ & 0.9407 & 0.9594 & 0.9074 & 0.9349  & 0.9356 \\ 
$\alpha=0.5$ & 0.9410 & 0.9574 & 0.9070 & 0.9344  & 0.9350 \\ 
 \hline
\end{tabular}
}

\label{Table_hyperparameter}
%\end{center}

\end{wraptable}

We investigate the effects of hyperparameter $\alpha$ under different levels of model quality, as shown in Tab. \ref{Table_hyperparameter}. We varied the hyperparameters $\alpha$ from 0.1 to 0.5, with 0.5 representing an equal mix of old and new features. We observed that when the quality of the old model is relatively high, the choice of hyperparameters $\alpha$ has minimal effect on the performance because of the good alignment between the old and new model. Conversely, when the quality of the old model is low, the choice of hyperparameters $\alpha$ can result in some performance fluctuations. Nonetheless, our method is still outperforms previous state-of-the-art approaches under various hyperparameter $\alpha$ settings.

Generally, the proportion of old features relative to new features should not be set too high, such as $\alpha=0.5$, since the new model is typically more robust and can further enhance retrieval performance. Conversely, setting the proportion too low, such as $\alpha=0.1$, may result in the new model not giving adequate attention to the old features. Therefore, we advocate for a moderate value, such as $\alpha=0.3$, to strike a balance between alignment and preserving optimization objectives.

\paragraph{The Effect of Denoising Operation.}

\begin{wraptable}{r}{7.3cm}
\vspace{-1.35em}
%\vspace{-1.5em}
\caption{The effect of denoising operation of old feature on performance under \textbf{different quality} of old model. "$\circleddash$" means no old feature denoising.}

\centering
\resizebox{\linewidth}{!}{
\begin{tabular}{lccccc}
\hline
\multicolumn{1}{l|}{}  & \multicolumn{2}{c|}{1:1 Verification} & \multicolumn{2}{c|}{1:N Identification}  \\ \cline{2-5}
\multicolumn{1}{l|}{}  & \multicolumn{1}{c|}{CT} & \multicolumn{1}{c|}{ST} & \multicolumn{1}{c|}{CT} & \multicolumn{1}{c|}{ST}  \\ \cline{2-5}
\multicolumn{1}{l|}{\multirow{-3}{*}{Method}} & \multicolumn{2}{c|}{TAR(\%)@FAR=10-4} & \multicolumn{2}{c|}{TPIR(\%)@FAR=10-2}  & \multicolumn{1}{c}{\multirow{-3}{*}{AVG}} \\ \hline

\rowcolor[gray]{0.9}
\multicolumn{6}{c}{Open-Class  \qquad \qquad  128-Dim   \qquad \qquad  $\mathcal{L}_{CE}$ / $\mathcal{L}_{Arc}$} \\ \hline
$Lower$-$B$ & {-} & {\color[HTML]{FF0000} 0.6358} & {-} & {\color[HTML]{FF0000} 0.4280} & {-} \\
 
$Upper$-$B$  & {0} & {\color[HTML]{4472C4} 0.9622} & {0} & {\color[HTML]{4472C4} 0.9419} & {\color[HTML]{4472C4} -}  \\
$MixBCT^{\circleddash}$ & 0.8267 & 0.9518 & 0.6874 & 0.9230  & 0.8472 \\ 
$MixBCT$  & 0.8305 & 0.9525 & 0.6973 & 0.9224 & \textbf{0.8507} \\ 
\hline
\rowcolor[gray]{0.9}
\multicolumn{6}{c}{Open-Class  \qquad \qquad  512-Dim   \qquad \qquad   $\mathcal{L}_{Arc}$ / $\mathcal{L}_{Arc}$} \\ \hline
$Lower$-$B$  & {-} & {\color[HTML]{FF0000} 0.9250} & {-} & {\color[HTML]{FF0000} 0.8783} & {-}\\
$Upper$-$B$  & {0.0002} & {\color[HTML]{4472C4} 0.9647} & {0} & {\color[HTML]{4472C4} 0.9453} & {\color[HTML]{4472C4} -}\\
$MixBCT^{\circleddash}$ & 0.9414 & 0.9598 & 0.9081 & 0.9372 & \textbf{0.9366} \\
$MixBCT$  & 0.9410 & 0.9597 & 0.9071 & 0.9367 & 0.9361  \\
 \hline
\end{tabular}
}

\label{Table_denoise}
\vspace{-1.35em}
\end{wraptable}

We also investigate the impact of denoising operation under different levels of model quality, as presented in Tab. \ref{Table_denoise}. The efficacy of denoising operation depends on the quality of the old model. When the quality of the old model is low, denoising can effectively reduce noise and facilitate backward-compatible training of the new model. However, for the relatively high-quality old model, denoising is no longer necessary due to the compactness of the old feature intra-class.

\section{Conclusion}
In this paper, we propose MixBCT, a simple yet effective backward-compatible training method that uses the new classifier to classify both old and new mixed features. In comparison to other state-of-the-art methods, MixBCT enables the new model to acquire distribution knowledge of old features and automatically adjust constraints. The training process of our method is straightforward and requires only a single classification loss function. We evaluate MixBCT on old models of varying quality and in multiple backward-compatible scenarios, demonstrating its effectiveness through extensive experiments.

\newpage
\appendix
\definecolor{OliveGreen}{rgb}{0.5, 0.65, 0.3}

\hrule height 4pt
\vskip 0.25in
\vskip -\parskip

\section*{\centering \fontsize{18}{22}\selectfont Appendix for MixBCT: Towards Self-Adapting Backward-Compatible Training}

\vskip 0.29in
\vskip -\parskip
\hrule height 1pt
\vskip 0.09in

\vspace{70pt}

\section{Summary}

This appendix provides visualizations, additional experiments, and pseudocode to further elaborate on and discuss our work.

\section{Visualization}

\subsection{Performance Comparison Visualization}

\begin{figure}[htbp]
\centering 
\includegraphics[width=0.7\textwidth]{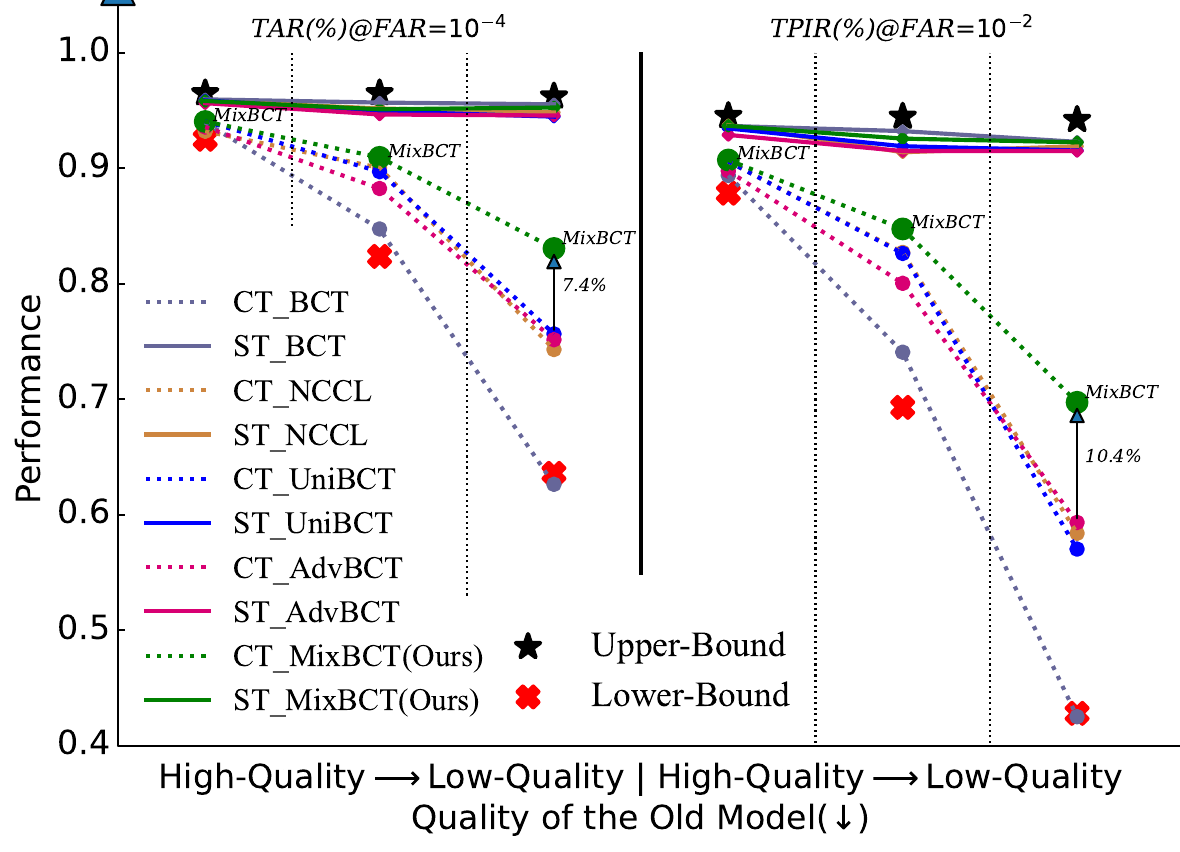}
\caption{Open-Class scenario, compare the 1:1 verification(left) and 1:N identification(right) performance on IJB-C face recognition benchmark with state-of-the-art methods in the setting of old model with different quality.  ‘CT' denotes ‘cross-test', measuring backward-compatible performance, while ‘ST' stands for ‘self-test', measuring the negative impact of backward-compatible training on new models. The x-axis on the graph represents the model quality, where points located further to the right indicate lower quality of the old model and higher within-class variance of the old embedding.}

\label{first_pic}
\end{figure}

In the ‘Open-Class' scenario, we offer a visual performance comparison in Fig. \ref{first_pic}, contrasting with Tab. \ref{Tabel_OC} in the main text. This comparison aims to more clearly demonstrate the superiority of our approach compared to other backward-compatible methods.

\subsection{Results t-SNE Visualization}

{For an old class, we conduct visualizations on it as well as on the three most closely related old classes. Subsequently, we showcase the comparative outcomes of four methods: UniBCT\cite{zhang2022uniBCT}, NCCL\cite{wu2022NCCL} and  AdvBCT\cite{pan2023boundary}, and our proposed approach, MixBCT. We present three illustrative examples, and their corresponding results are displayed in Fig. \ref{fig:feature}. We can observed  that MixBCT yields better results in achieving the desired objective (as defined in Eq. 2 of the main text) of backward-compatible training.

\begin{figure*}[htbp]
    \centering 
    \captionsetup[subfigure]{labelformat=empty}
    \setlength{\abovecaptionskip}{-1pt}
    \setlength{\belowcaptionskip}{2pt}
    \subfloat[]{\includegraphics[width=0.32\textwidth]{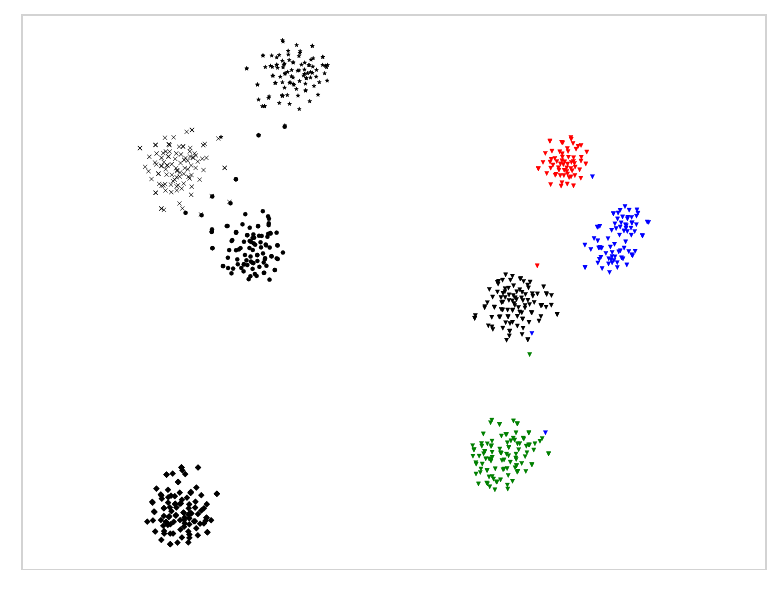}}
    \subfloat[]{\includegraphics
         [width=0.32\textwidth]{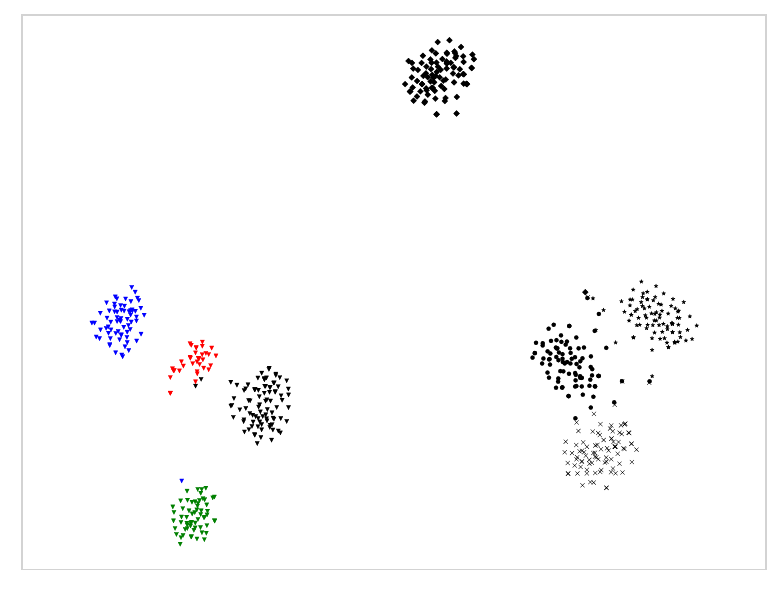}}
    \subfloat[]{\includegraphics
         [width=0.32\textwidth]{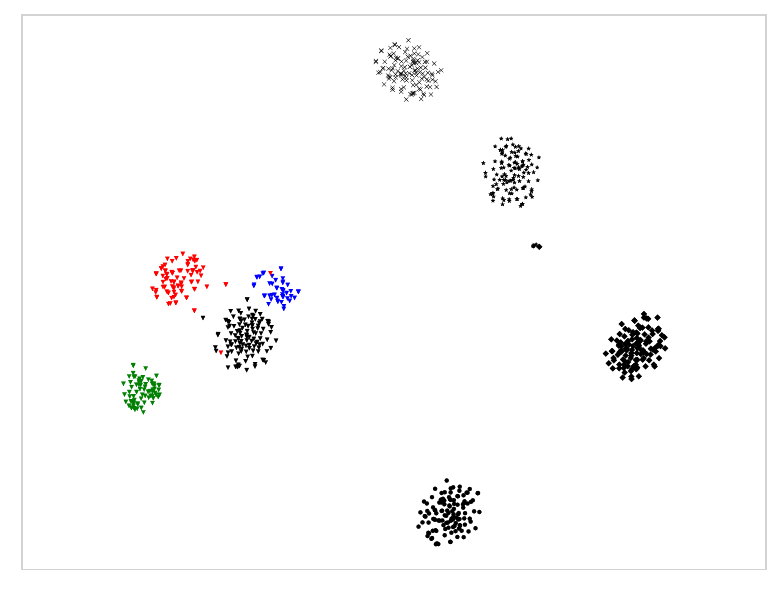}}

    \caption{Feature visualization with different colors for different classes when the old model has low quality, where old feature can't separate well. ‘$\blacktriangledown$': Old, ‘$\bigstar$': NCCL, ‘$\times$': UniBCT, ‘$\blacksquare$': AdvBCT ,‘$\bullet$': MixBCT}
    \label{fig:feature}
    \vspace{-5mm}
\end{figure*}

\subsection{Denoise t-SNE Visualization}

We demonstrated the visualization results of the denoising operation in Fig. \ref{noise} under the relatively low quality of the old model. As observed, the denoising operation effectively removes noise from the old features.

\begin{figure}[htbp]
\centering 
\includegraphics[width=0.45\textwidth]{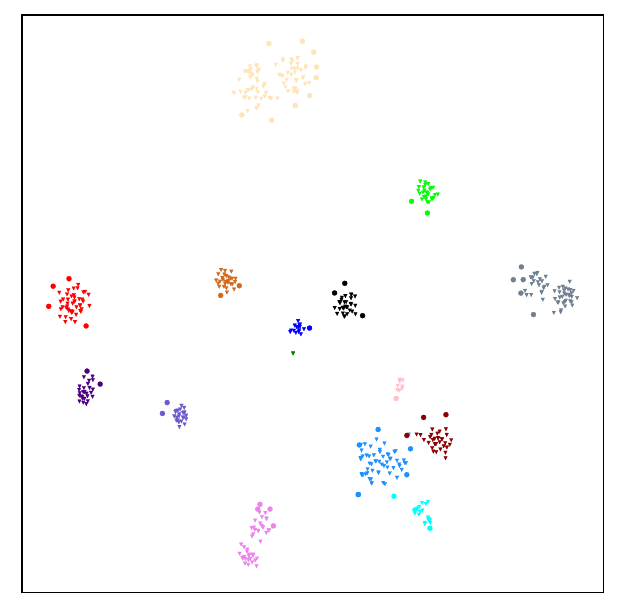}
\caption{Denoising operation visualizations, different colors represent different categories, and ‘$\bullet$' represents noisy samples.}

\label{noise}
\end{figure}

\section{Additional Experiments}
We conducted additional experiments to further validate the superiority of our method. Similar to the main text, we compared it with L2 baseline and current representative 
 state-of-the-art methods: BCT\cite{shen2020BCT}, UniBCT\cite{zhang2022uniBCT}, NCCL\cite{wu2022NCCL} and  AdvBCT\cite{pan2023boundary}.

\subsection{More Results about the Comparison with Another Tuned L2 Regression Baseline}

As demonstrated in the main text, when the quality of the old model is low, L2 regression with carefully adjusted constraint strength is a competitive baseline. To further highlight the superiority of our method, we conducted comparisons with another sufficiently tuned L2 regression baseline, which is based on denoised old features (utilized in MixBCT). We present the results under multiple settings for the hyperparameter $\lambda$, constructing multiple levels of constraint strengths. As shown in Tab. \ref{Tabel_tuned_L2}, it can be observed that our method still maintains a significant advantage.

\begin{table}[htbp]
\renewcommand{\arraystretch}{1.1}
\begin{center}
\caption{Comparison between MixBCT and another tuned L2 loss.}
\label{Tabel_tuned_L2}
%\resizebox{0.7\linewidth}{!}{
\begin{tabular}{lcccccc}
\hline
\multicolumn{2}{l|}{} &
  \multicolumn{2}{c|}{1:1 Verification} &
  \multicolumn{2}{c|}{1:N Identification} &
   \\ \cline{3-6}
 
  & & \multicolumn{1}{|c|}{CT} &
  \multicolumn{1}{c|}{ST} &
  \multicolumn{1}{c|}{CT} &
  \multicolumn{1}{c|}{ST} &
  \\ \cline{3-6}
\multicolumn{2}{c|}{\multirow{-3}{*}{Setup}} & \multicolumn{2}{c|}{TAR@FAR=10-4} & \multicolumn{2}{c|}{TPIR@FAR=10-2}  & \multicolumn{1}{c}{\multirow{-3}{*}{AVG}}  \\ \hline

\rowcolor[gray]{0.9}
\multicolumn{7}{c}{Open-Class  \qquad \qquad  128-Dim    \qquad \qquad  $\mathcal{L}_{CE}$ / $\mathcal{L}_{Arc}$}   \\ 
\hline

\multicolumn{2}{l}{${L_2}^{*}$, \quad  $\lambda=1$}  & 0.5810 & 0.9548 & 0.4051 & 0.9298  & 0.7177 \\
\multicolumn{2}{l}{${L_2}^{*}$, \quad $\lambda=5$}  & 0.7580 & 0.9436 & 0.6158 & 0.9140  & 0.8079 \\
\multicolumn{2}{l}{${L_2}^{*}$, \quad  $\lambda=6$}  & 0.7800 & 0.9405 & 0.6547 & 0.9091  & 0.8211 \\
\multicolumn{2}{l}{${L_2}^{*}$, \quad   $\lambda=10$}  & 0.8085 & 0.9359 & 0.6925 & 0.9030  & \textbf{0.8350} \\
\multicolumn{2}{l}{${L_2}^{*}$, \quad  $\lambda=15$}  & 0.8133 & 0.9304 & 0.6978 & 0.8891  & 0.8327 \\
\multicolumn{2}{l}{${L_2}^{*}$, \quad   $\lambda=20$}  & 0.8088 & 0.9209 & 0.6708 & 0.8616  & 0.8155 \\
\hline
\multicolumn{2}{l}{MixBCT}  & 0.8305& 0.9525 & 0.6973 & 0.9224  & \textbf{0.8507} \\

\hline
\end{tabular}
%}

\end{center}
\end{table}
\vspace{-1em}

\subsection{More Results about of Backbone Changes}

In the experiments of mian text, we used iResnet18\cite{duta2021improved} to train the old model and iResnet50\cite{duta2021improved} to train the new model. In this section, we tested two additional combinations of network structures:

1) Both the old and new model were trained using the same network structure: iResnet18.

2) Old and new model trained on different network architectures, where the old model was trained using the CNN architecture: iResnet18, while the new model was trained using the transformer\cite{transformer} architecture: ViT-s\cite{vit}.

We also conducted experiments in the most challenging Open-Class scenarios, and the experimental results are shown in Tab. \ref{backbone_change}. It can be seen that our method achieved significant performance improvements in all settings.

\vspace{-1em}
\begin{table}[htbp]

\caption{More compatibility results of backbone changes.}
\label{backbone_change}
\centering
\begin{tabular}{lccccccc}
\hline

\multicolumn{1}{l|}{\multirow{3}{*}{Method}} &
\multicolumn{1}{c|}{\multirow{3}{*}{Old}} &
\multicolumn{1}{c|}{\multirow{3}{*}{New}} &
\multicolumn{2}{c|}{1:1 Verification} &
\multicolumn{2}{c|}{1:N Identification} &
\multirow{3}{*}{Avg} \\ \cline{4-7}

\multicolumn{1}{l|}{} &
  \multicolumn{1}{c|}{} &
  \multicolumn{1}{c|}{} &
  \multicolumn{1}{c|}{CT} &
  \multicolumn{1}{c|}{ST} &
  \multicolumn{1}{c|}{CT} &
  \multicolumn{1}{c|}{ST} &
   \\ \cline{4-7}
    \multicolumn{1}{l|}{} &
  \multicolumn{1}{c|}{} &
  \multicolumn{1}{c|}{} & \multicolumn{2}{|c|}{TAR@FAR=10-4} & \multicolumn{2}{c|}{TPIR@FAR=10-2}  &  \\ \hline
\rowcolor[gray]{0.9}
\multicolumn{8}{c}{Open-Class  \qquad \qquad  128-Dim    \qquad \qquad  $\mathcal{L}_{CE}$ / $\mathcal{L}_{Arc}$}   \\ 
\hline
$Lower$-$B$ & iR18 & & {\color[HTML]{FF0000} -} & {\color[HTML]{FF0000} 0.6358} & {\color[HTML]{FF0000} -} & {\color[HTML]{FF0000} 0.4280} & {\color[HTML]{FF0000} -} \\

$Upper$-$B$  & &  iR18 & {0} & {\color[HTML]{4472C4} 0.9412} & {0} & {\color[HTML]{4472C4} 0.9103} & {\color[HTML]{4472C4} -}  \\

$BCT$   & iR18 & iR18     &  0.6543 & 0.9408 &  0.4440 & 0.9123 & 0.7378 \\
$UniBCT$   & iR18 & iR18     & 0.7348 & 0.9288 & 0.5573 & 0.8911 & 0.7780 \\

$NCCL$   & iR18 & iR18 &  0.7233      &  0.9318     &   0.5294     &   0.8954     &   0.7700     \\
$AdvBCT$   & iR18 & iR18    & 0.7613 & 0.9356 & 0.6000 & 0.8992 & 0.7990 \\
$L_2$   & iR18 & iR18     & 0.7590 & 0.9113 & 0.5770 &  0.8368 & 0.7710 \\

$MixBCT$ & iR18 & iR18 & 0.7904 & 0.9315 & 0.6404 & 0.8893 & \textbf{0.8129} \\
\hline
$Lower$-$B$ & iR18 & & {-} & {\color[HTML]{FF0000} 0.6358} & {-} & {\color[HTML]{FF0000} 0.4280} & {-} \\
$Upper$-$B$  & &  ViT-s & { 0.0003} & {\color[HTML]{4472C4} 0.9253 } & {0} & {\color[HTML]{4472C4} 0.8731} & {\color[HTML]{4472C4} -}  \\
$BCT$   & iR18 & ViT-s     &  0.5906 & 0.9384 &  0.3687 & 0.8937 & 0.6978 \\
$UniBCT$   & iR18 & ViT-s     & 0.6844 & 0.9370 & 0.4840 & 0.8925 & 0.7495 \\
$NCCL$   & iR18 & ViT-s     & 0.6572 & 0.9259 & 0.4255 & 0.8688 & 0.7194 \\
$AdvBCT$   & iR18 & ViT-s     & 0.6779 & 0.9368 & 0.4901 & 0.8897 & 0.7486 \\
$L_2$   & iR18 & ViT-s     & 0.7154 & 0.9231 & 0.5128 &  0.8459&  0.7493 \\

$MixBCT$ & iR18 & ViT-s     & 0.7939 & 0.9396 & 0.6320 & 0.9000 & \textbf{0.8164} \\
\hline
\end{tabular}

\end{table}

\subsection{Additional Experiments on Person Re-ID Task}

Face recognition is an important scenario for backward-compatible training, so we conducted experiments on face datasets and validated the results in 9 cases: the quality of old models is from low to high, and four different setups. Here, we performed an additional experiment on the challenging Open-Class scenario using the Market1501\cite{Market1501} dataset for person re-id task.

Market1501 is a widely used benchmark dataset for person re-identification tasks. It consists of 1501 pedestrians captured by 6 cameras in the campus of Tsinghua University, with a total of 32,668 annotated bounding boxes. The dataset comprises 12,936 images of 751 pedestrians for the training set, and 19,732 images of the remaining 750 pedestrians for the testing set. During testing, 3,368 manually annotated images of 750 pedestrians are used as the query set, while the remaining images constitute the gallery set. The gallery set is automatically detected using the Deformable Part Model (DPM) based detector\cite{dpm} .

We utilized Resnet50 as the backbone network. And the old model was trained on 30\% of available classes with the Cross-Entropy loss, while the new model was trained on the remaining 70\% of classes with both Cross-Entropy loss and Triplet loss\cite{schroff2015facenet}. The results are presented in Tab. \ref{reid}. It can be observed that our method still exhibits significant performance advantages.

\begin{table}[htbp]
\caption{Comparison on the person re-id task.}
\label{reid}

\renewcommand{\arraystretch}{1.2}
\resizebox{\linewidth}{!}{
\begin{tabular}{lcccccccc}

\hline
\multicolumn{1}{l|}{\multirow{2}{*}{Method}} & \multicolumn{4}{c|}{ST}                & \multicolumn{4}{c}{CT} \\ \cline{2-9} 
\multicolumn{1}{c|}{}            & Rank-1  & Rank-5  &  Rank-10 & \multicolumn{1}{c|}{mAP}    & Rank-1  & Rank-5  & Rank-10 & mAP \\ \hline
\rowcolor[gray]{0.9}
\multicolumn{9}{c}{Open-Class  \qquad \qquad  2048-Dim    \qquad \qquad  $\mathcal{L}_{CE}$ / $\mathcal{L}_{CE+Triplet}$}   \\ 
\hline
\multicolumn{1}{l|}{$Lower$-$B$} & \color[HTML]{FF0000}0.8560 & \color[HTML]{FF0000}0.9442 & \color[HTML]{FF0000}0.9635 &\multicolumn{1}{c|}{\color[HTML]{FF0000}0.6828} & -     & -   & -  & -   \\ 
\multicolumn{1}{l|}{$Upper$-$B$} & \color[HTML]{4472C4}0.9320 & \color[HTML]{4472C4}0.9765 & \color[HTML]{4472C4}0.9872 &\multicolumn{1}{c|}{\color[HTML]{4472C4}0.8212} & 0.2556     & 0.4620    & 0.5585 & 0.1483   \\ 
\multicolumn{1}{l|}{$BCT$} &   0.9145     &   0.9718       & 0.9831  & \multicolumn{1}{c|}{0.8086 }   & 0.8450   & 0.9469    &  0.9662  &     0.6837   \\ 
\multicolumn{1}{l|}{$UniBCT$}  & 0.9186    &  0.9718      &   0.9825     & \multicolumn{1}{c|}{0.8097}       &  0.3005     &   0.5205    &  0.6185 &  0.1860   \\ 

\multicolumn{1}{l|}{$NCCL$}       & 0.8705 & 0.9418 & 0.9638 & \multicolumn{1}{c|}{0.7062} & 0.7268   & 0.8818   & 0.9270  & 0.5601   \\ 

\multicolumn{1}{l|}{$AdvBCT$}      & 0.8527 &   0.9329      & 0.9564 &   \multicolumn{1}{c|}{0.6807}       &   0.7221    &    0.8786    & 0.9228 &   0.5509  \\ 

\multicolumn{1}{l|}{$L_2$}                      & 0.8999 & 0.9620  & 0.9754 & \multicolumn{1}{c|}{0.7683} & 0.8536   & 0.9480   & 0.9679  & 0.6839   \\

\multicolumn{1}{l|}{$MixBCT$}                  & \textbf{0.9276} & \textbf{0.9762}  & \textbf{0.9857} & \multicolumn{1}{c|}{\textbf{0.8149}} & \textbf{0.8884}   & \textbf{0.9605}   & \textbf{0.9780}  & \textbf{0.7314}   \\ \hline
\end{tabular}

}

\end{table}

\subsection{Multi-model and Sequential Compatibility}
\label{Sequential}

In this section, we showcase the capability of our method in multi-model sequence updating. Consistent with the main text, we also conducted experiments based on MS1Mv3 \cite{deng2019arcface} and IJB-C \cite{maze2018iarpa}. We trained three models: ${\phi}_{1}$, ${\phi}_2$, and ${\phi}_3$, with each subsequent model being trained on the foundation of the previous one. We configured the embedding dimension to 128. The experimental setups for these three models are delineated in Tab. \ref{setting_detail}. Specifically, they utilized iResnet18, iResnet18, and iResnet50 as the backbone networks, employing Cross-Entropy loss, Arcface\cite{deng2019arcface} loss and Arcface loss. The training sets comprised the top 30\%, 70\%, and 100\% classes of the MS1Mv3.

\begin{table}[htbp]

\centering
\caption{The experimental setup details of the three models: ${\phi}_{1}$, ${\phi}_2$, and ${\phi}_3$}
\begin{tabular}{c|c|c|c|c|c|c}

\hline
Model & New \# Backbone & Old  \#  Backbone  & Loss & Traing Data & Classes & Images \\ \hline
${\phi}_1$  & \quad ${\phi}_1$  \#  iR18  &  -   \#  -  & $L_{CE}$ & Top 30\%-Class & 28,029 & 1,581,241 \\
${\phi}_2$  & \quad ${\phi}_2$ \# iR18  &  \quad ${\phi}_1$   \# iR18  &  $L_{Arc}$ & Top 70\%-Class & 65,402 & 3,619,758 \\

${\phi}_3$  & \quad ${\phi}_3$   \# iR50 &  \quad ${\phi}_2$  \#  iR18  & $L_{Arc}$ & 100\%-Class & 93,431 & 5,179,510 \\ \hline
\end{tabular}
\label{setting_detail}
\end{table}

We conducted evaluations on both 1:1 face verification and 1:N face identification protocols. The multi-model and sequential compatibility results of our method are illustrated in Fig. \ref{Sequential_Comp}. As shown, although ${\phi}_3$ did not directly use the information from ${\phi}_1$ for backward-compatible training, it still exhibits good backward compatibility with ${\phi}_1$, even outperforming ${\phi}_2$, which was directly trained with ${\phi}_1$ for backward compatibility.

\begin{figure}[htbp]
\centering 
\includegraphics[width=0.7\textwidth]{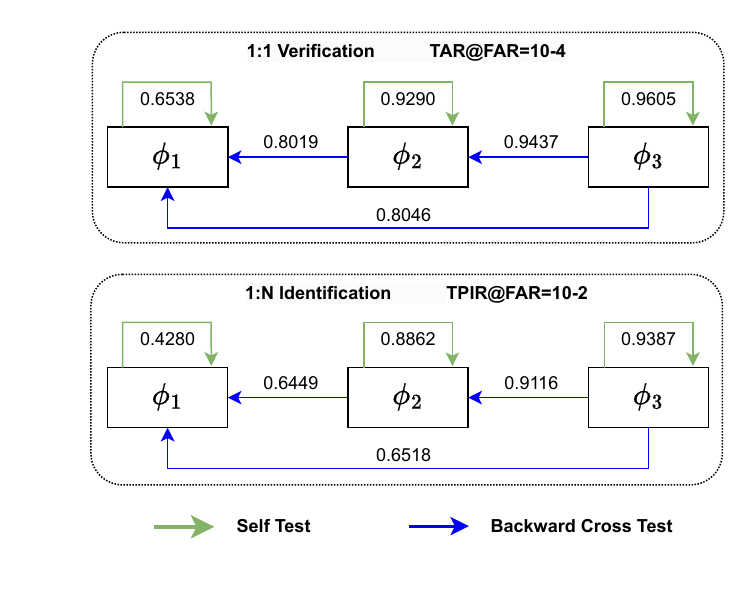}
\caption{Multi-model and Sequential Compatibility Results }

\label{Sequential_Comp}
\end{figure}

\section{Pseudo Training Code}

{\normalsize We illustrate the core algorithm of MixBCT in Alg. \ref{alg:MixBCT}}
\begin{algorithm}[htbp]\large
    \caption{Pseudocode of MixBCT in a PyTorch-like style.} 
    \label{alg:MixBCT} 
    
    \normalsize
    \textcolor{OliveGreen}{\#old\_features: $\phi_o(\mathcal{D}_n)$} \\ 
    \textcolor{OliveGreen}{\#old\_cred\_tag: credible tags of old feature; \ if credible:1 }\\
    \textcolor{OliveGreen}{\#new\_model: the backbone of the new mode} \\
    \textcolor{OliveGreen}{\#bs: batch size} \\
    \textcolor{OliveGreen}{\#ratio: mix ratio} \\
    for ep in range(start\_ep, end\_ep): \\
    \mbox{}\quad for img, label, index in train\_loader: \\
    \mbox{}\qquad  \textcolor{OliveGreen}{\#get selceted index} \\
    \mbox{}\qquad  bs\_cred = old\_cred\_tag[index] \\ 
    \mbox{}\qquad  bs\_cred = torch.where(bs\_cred == 1)[0] \\ 
    \mbox{}\qquad  sel\_ind = random.sample(range(0, len(bs\_cred)),int(bs * ratio)) \\
    \mbox{}\qquad  sel\_ind = bs\_cred[sel\_ind] \\ 
    \mbox{}\qquad \textcolor{OliveGreen}{\#chose old features} \\
    \mbox{}\qquad  old\_emb = old\_features[index]\\
    \mbox{}\qquad  new\_emb = new\_model(img) \\
    \mbox{}\qquad \textcolor{OliveGreen}{\#mix operation} \\
    \mbox{}\qquad new\_emb[sel\_ind] = old\_emb[sel\_ind] \\ 
    \mbox{}\qquad \textcolor{OliveGreen}{\#calculate loss use the combined features} \\
    \mbox{}\qquad loss = $L_{cls}$(new\_emb, label) \\ 
    \mbox{}\qquad \textcolor{OliveGreen}{\#loss backward} \\
    \mbox{}\qquad loss.backward() \\
    \mbox{}\qquad update(new\_model.params) \\ 
\end{algorithm}%\small

\section{Limitations and Future Work}
This work focuses on designing a backward-compatible training framework that is simpler, easier to implement, more broadly applicable, and performs better than existing methods. Our approach involves a hyperparameter $\alpha$. Although we have provided some discussions on the selection of $\alpha$ in Section 4.4 of the main text, online adjustment or adaptive tuning of $\alpha$ is interesting and has the potential to improve performance. This will be part of our future work.

%%%%%%%%%%%%%%%%%%%%%%%%%%%%%%%%%%%%%%%%%%%%%%%%%%%%%%%%%%%%
\newpage
\bibliographystyle{plain}

\end{document}